\documentclass[10pt,twocolumn,letterpaper]{article}

\usepackage{wacv}
\usepackage{times}
\usepackage{epsfig}
\usepackage{graphicx}
\usepackage{amsmath}
\usepackage{amssymb}
\usepackage[ruled,vlined]{algorithm2e}
\usepackage{multirow}
\usepackage{graphicx}
\usepackage{subcaption}
\usepackage{wrapfig}
\usepackage{pifont}
\usepackage{threeparttable}
\usepackage{wrapfig,lipsum,booktabs}
\usepackage{xcolor, colortbl}
\usepackage{color}
\usepackage{verbatim}
\usepackage{float}
\usepackage{enumitem}

\def\wacvPaperID{522} 

\wacvfinalcopy 
\wacvalgorithmstrack
\ifwacvfinal
\def\assignedStartPage{1} 
\fi

\ifwacvfinal
\usepackage[breaklinks=true,bookmarks=false]{hyperref}
\else
\usepackage[pagebackref=true,breaklinks=true,colorlinks,bookmarks=false]{hyperref}
\fi

\ifwacvfinal
\else
\fi

\input{def}

\makeatletter
\newcommand{\removelatexerror}{\let\@latex@error\@gobble}
\makeatother

\begin{document}

\title{ PointACL:Adversarial Contrastive Learning for Robust Point Clouds Representation under Adversarial Attack}

\author{Junxuan Huang\\
University at Buffalo\\
{\tt\small Junxuanh@buffalo.edu}
\and
Yatong An\\
Xmotors.ai\\
{\tt\small yatong.an@gmail.com}
\and
Cheng Lu\\
Xmotors.ai\\
{\tt\small luc@xiaopeng.com}
\and
Bai Chen\\
Xmotors.ai\\
{\tt\small chenbai@xiaopeng.com}
\and
Junsong Yuan\\
University at Buffalo\\
{\tt\small jsyuan@buffalo.edu}
\and
Chunming Qiao\\
University at Buffalo\\
{\tt\small qiao@buffalo.edu}
}

\maketitle

\begin{abstract}
    Despite recent success of self-supervised based contrastive learning model for 3D point clouds representation, the adversarial robustness of such pre-trained models raised concerns. Adversarial contrastive learning (ACL) is considered an effective way to improve the robustness of pre-trained models. In contrastive learning, the projector is considered an effective component for removing unnecessary feature information during contrastive pretraining and most ACL works also use contrastive loss with “projected” feature representations to generate adversarial examples in pretraining, while "unprojected " feature representations are used in generating adversarial inputs during inference. Because of the distribution gap between “projected” and "unprojected" features, their models are constrained of obtaining robust feature representations for downstream tasks. We introduce a new method to generate high-quality 3D adversarial examples for adversarial training by utilizing virtual adversarial loss with “unprojected” feature representations in contrastive learning framework. We present our robust aware loss function to train self-supervised contrastive learning framework adversarially. Furthermore, we find selecting high difference points with the Difference of Normal (DoN) operator as additional input for adversarial self-supervised contrastive learning can significantly improve the adversarial robustness of the pre-trained model. We validate our method, \textsc{PointACL} on downstream tasks, including 3D classification and 3D segmentation with multiple datasets. It obtains comparable robust accuracy over state-of-the-art contrastive adversarial learning methods.

\end{abstract}

\section{Introduction}

Among various 3D representation methods, point clouds are popular for scene understanding and visual analysis. Tasks include 3D object classification, detection, and segmentation. Despite its popularity, adversarial robustness of its learned 3D perception models, namely, robustness against adversarial samples, is a major security concern in real-world application.
Perturbed adversarial samples like point adding\cite{xiang2019generating}, point dropping\cite{zheng2018pointcloud} or point shifting\cite{extend_liu} can easily mislead 3D perception models.

Adversarial training (AT) \cite{AT2017} and its variants\cite{chen2019robust,boopathy2020proper,shafahi2019adversarial} are considered effective defense strategies against adversarial attacks. However, they rely on class labels to generate adversarial samples that are used to supervise model training for robustness and it is difficult to conduct  unsupervised training methods like self-supervised learning. Sun \etal~\cite{sun2021adversarially} proposed a pretext tasks based self-supervised (SSL) adversarial training method, which forces the model to learn robust representation from solving a pre-designed pretext task without class labels.
Some previous works like \cite{IFD} build an additional networks connected to classifier as 3D point clouds purifier. 
Researchers later introduced contrastive learning framework as an extension to AT and SimCLR\cite{simclr}. It  projects feature representations into a different dimension space for conducting contrastive loss and the contrastive loss is also used to obtain adversarial examples of 2D images without using class labels \cite{kim2020adversarial,acl,advcl}. But their contrastive loss guided adversarial examples rely on projected feature in contrastive learning framework, while unprojected feature are used in generating adversarial inputs in downstream task. Because of the distribution gap between unprojected feature and projected feature, contrastive loss guided adversarial 
examples can only provide limited robustness during adversarial pretraining stage.

In this work, we designed an adversarial contrastive learning model specially for 3D point clouds, \textsc{PointACL}. Specifically, in order to mitigate the distribution gap between training examples and testing adversarial inputs, we propose a novel method for generating adversarial examples with unprojected features. We introduce virtual adversarial loss\cite{miyato2018virtual} on the unprojected features to calculate gradient direction for perturbation, which is better than prior adversarial contrastive learning methods\cite{kim2020adversarial,acl,advcl}. Leveraging the virtual adversarial loss, our method lowers the divergence between adversarial samples and perturbed adversarial inputs in testing, and experiments show that our model has an advantage in downstream tasks under adversarial robustness testing.
Meanwhile, to enhance the robustness of the feature representation, we choose to add normal of point clouds surface information in adversarial pretraining. We first select point clouds from Difference of Normal(DoN) operator with information on surface gradient and treat those high difference points as additional input in pretraining. During adversarial contrastive learning, we extracted projected representations of high difference points and incorporated them into multi-view contrastive loss.


To verify the advantages of our proposed network, we applied our pre-trained network with standard linear finetuning on two downstream tasks : 3D classification and 3D segmentation. Specifically, for the classification network, we trained on ModelNet and tested on ModelNet and shapeNet. The robust accuracy on ModelNet achieved \textbf{27.51\%} compared to \textbf{4.03\%} without adversarial training and in shapeNet we achieved \textbf{13.34\%} compared to \textbf{2.13\%} without adversarial training. For 3D segmentation task, robust accuracy achieved was \textbf{39.08\%} compared to \textbf{13.82\%} without adversarial training.

Our major contributions can be summarized as follows:

\ding{182} We proposed \textsc{PointACL}, an adversarial contrastive learning framework for point clouds data, and we found that using the virtual adversarial loss to generate high-quality adversarial samples during pre-training stage can bring more robustness representation in downstream tasks.

\ding{183} We verified that high-difference point clouds selected from the difference of normal (DoN) operator can contribute to the robustness of 3D representation learning. Experiments on downstream tasks verified the criticalness of high-difference points in improving the network's adversarial robustness. 

\ding{184} We extensively benchmarked our pre-trained model with other adversarial contrastive learning models. Our pre-trained models tested on two downstream tasks : \textit{3D object classification} (on ModelNet40) and \textit{3D segmentation} (on S3DIS) under standard linear finetuning  \textsc{PointACL} led to new improved state-of-the-art robust accuracy. For example, in the 3D object classification task with I-FGM attack under budget with $\epsilon$=0.01m in 7 steps we achieve {\textbf{18.72\%} and  \textbf{17.24\%}} robustness improvement over existing adversarial contrastive learning methods\cite{kim2020adversarial} and \cite{acl}. 

\section{Related works}
In this section, we overview the progress on three related topics: adversarial attacking methods for point clouds, self-supervised Learning of point clouds and defensive methods in improving adversarial robustness.
\subsection{Adversarial attack on Point Clouds}
The robustness of deep learning model on 3D point clouds has attracted many researchers due to its applications in
robotics and safe-driving cars. Existing 3D adversarial attack methods can be roughly divided into three classes: optimization-based, gradient-based, and generation-based. For gradient-based methods, Liu \etal~\cite{extend_liu} extended I-FGSM\cite{IFGM} into the 3D point clouds domain by perturbing the point coordinates. Zheng \etal~\cite{zheng2018pointcloud} proposed an iterative point dropping attack by building a gradient-based saliency map. In optimization-based methods, xiang\etal~\cite{xiang2019generating} first proposed to generate adversarial point clouds using C\&W attack framework\cite{CW} by point perturbation and adding. However,those perturbed point clouds usually contain many
outliers, which are not human-unnoticeable. 
To reduce those outliers, Wen \etal~\cite{wen2020} focus on generating adversarial point cloud with much less outliers.
LG-GAN \cite{LG-GAN} is a generation-based 3D attack method, which uses GANs\cite{goodfellow2014generative} to generate adversarial point clouds that follows the input target labels. 
\subsection{Self-supervised Learning of Point Clouds}
Many approaches \cite{wu2016learning,achlioptas2017representation,gadelha2018multiresolution,yang2018foldingnet} have been proposed for unsupervised learning and generation of point clouds, but high-level downstream tasks like 3D object classification and segmentation is less discussed. Some recent work try to demonstrate the potentials for high-level tasks such as 3D object classification and 3D semantic segmentation. Those self-supervised learning methods can be divided into two classes: pretext task based approach and contrastive learning based approach.
Poursaeed \etal~\cite{SSL_Orientation} designed a pretext task that predicts rotations angle of point clouds object and achieved good performance in 3D object classification.
Compare to pretext task based approach, contrastive learning methods have advantage in transferability and generalisability, they have achieved great success in 2D image domain. For 3D point clouds domain, Xie \etal~\cite{pointcontrast} used contrastive loss between two different transformations results of point clouds.
STRL\cite{huang2021spatio} extended BYOL\cite{grill2020bootstrap} structure to point clouds domain by utilizing the spatio-temporal contexts and structures of point clouds and it achieved state-of-the-art performance in various high-level downstream tasks.

\begin{figure*}[htp]
	\centering
	\includegraphics[width=0.9\linewidth]{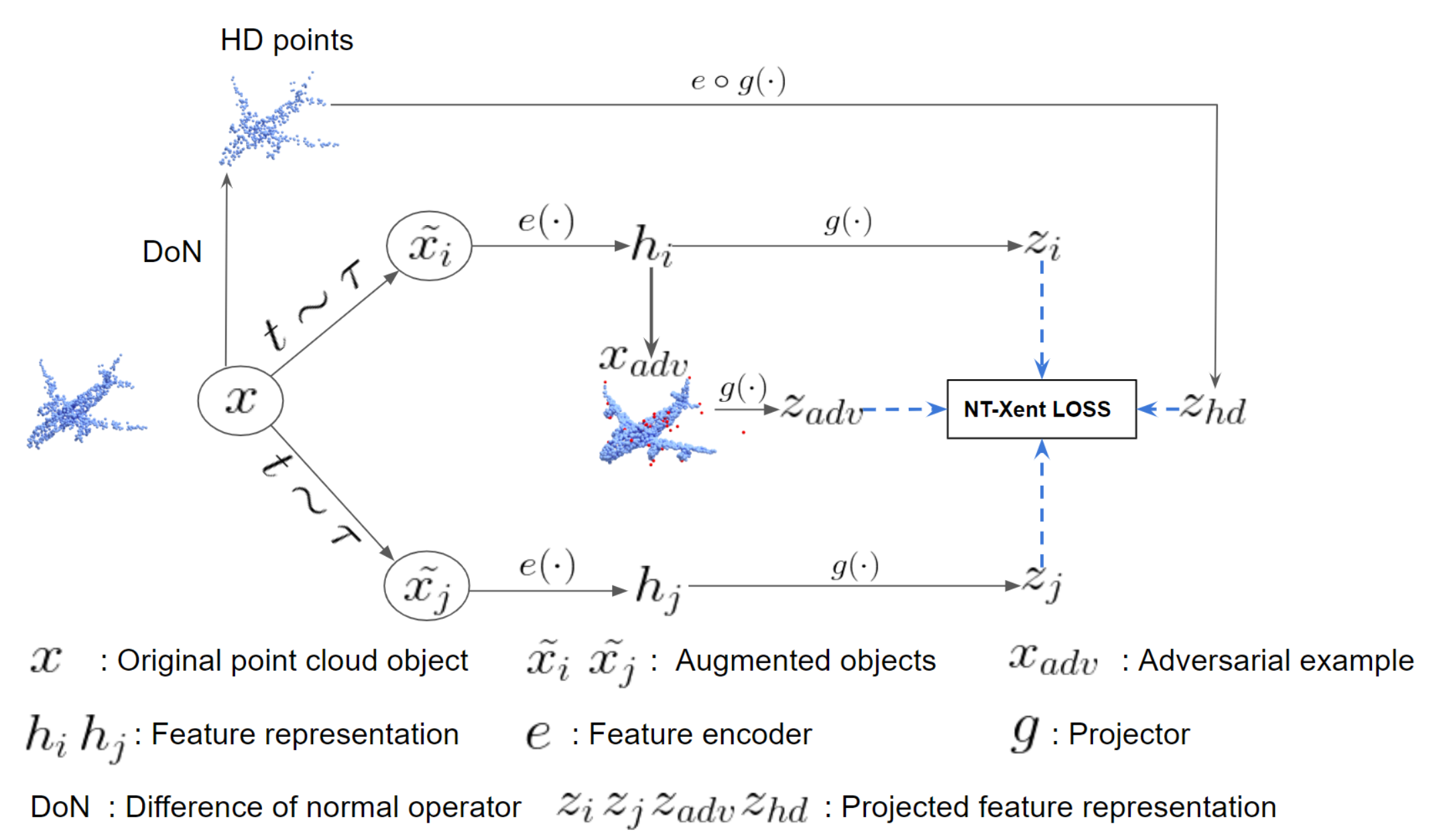}
	\caption{\textbf{Overview of PointACL.}
	${x}$: Input point clouds object,
	At first, the input point clouds object ${x}$ will generate three different versions of inputs: $\tilde{x}_{i},\tilde{x}_{j},x_{hd}$, corresponding to two augmented objects and high-difference point object. Adversarial examples will be generated based on one of the augmented object. Same as SimCLR\cite{simclr}, we have a mlp based nonlinear projection head $g$ to improve representation quality and a multi-view normalized cross entropy loss with adjustable temperature (NT-Xent\cite{simclr}) will be optimized based on projected feature representations $z$ as the contrastive loss.}
	\label{fig:overview}
\end{figure*}
\subsection{Adversarial robustness}
Many defense methods have been proposed to improve model robustness against adversarial attacks. Adversarial training(AT)\cite{AT2017}  provides one of the most effective defense methods by training the model over the adversarially perturbed
training data. 
Zhou \etal~\cite{dup-net} used Statistical Outlier Removal (SOR) method for removing points with a large kNN distance. They also proposed DUPNet, which is a combination of SOR and a point cloud upsampling network PU-Net\cite{yu2018punet}.In contrastive learning, RoCL\cite{kim2020adversarial} and ACL\cite{acl} AdvCL\cite{advcl} extended AT by using contrastive loss to eliminate the need for class labels when generating adversarial samples. VAT\cite{miyato2018virtual} proposed a new regularization method based loss to generate adversarial samples on unlabeled data in Semi-Supervised Learning.

\label{headings}
\section{Our approach}
\subsection{ Definition and Notations}
\paragraph{Point Cloud.} 

For classification task, a point cloud object is represented as $(\xb \triangleq \{\pb_i\}_{i = 1...N},  y)$, where $\pb_i \in \mathbb{R}^3$ is a 3D point and $N$ is the number of points in the point cloud; $y \in \{1, 2,...,k\}$ is the ground-truth label, where $k$ is the number of classes. For semantic segmentation, a point cloud is $(\xb \triangleq \{\pb_i\}_{i = 1...N},  \{\yb_i\}_{i = 1...N})$, which $\yb_i \in \{1, 2,...,k\}$ represents ground-truth label for each point.

\paragraph { Contrastive learning.}  
 The main idea of contrastive learning is to learn representations without supervision by maximizing agreements of different augmented views of the same point cloud. Inspired by SimCLR\cite{simclr} and STRL\cite{huang2021spatio}, we build our 3D contrasive learning framework. Specifically, consider augmentations $\mathcal{T}$ (combination of rotation, translation, scaling, cropping, cutout,jittering and down-sampling) in Fig.\ref{fig:overview}, each unlabeled point cloud ${x}$ is augmented as $\tilde{x}_{i}$ and $\tilde{x}_{j}$. The feature encoder generates output feature $({\boldsymbol{h}}_{i},{\boldsymbol{h}}_{j})$ from pair $(\tilde{x}_{i},\tilde{x}_{j})$. Then, MLP-based projector is applied for  feature extraction. The projected feature representations $({\boldsymbol{z}}_{i},{\boldsymbol{z}}_{j})$ are optimized under a contrastive loss $\ell_{NT}$(NT-Xent) to maximize their agreement. After training, we keep the the encoder part as the pre-trained model for downstream tasks.

\paragraph{Unsupervised Adversarial Training.}
Different from adversarial training (AT) which is supervised learning, Unsupervised Adversarial Training (UAT) does not use labels to generate adversarial samples. Compared with other unsupervised adversarial training algorithms, ours improves robustness of 3D point cloud (self-supervised) pretraining model in an unsupervised manner.

\subsection{ Problem statement}
Our method aims to improve the robustness of the 3D self-supervised contrastive learning pretraining model. Following \cite{AT2017}, we add adversarial examples during the pretraining stage and aim to minimize contrastive loss between adversarial examples and normal inputs. So that we could achieve robust feature representations in the pretraining stage and, after linear finetuning, make more accurate predictions under adversarial attacks, improving the model's adversarial robustness performance on downstream tasks such as 3D classification and 3D semantic segmentation. We can formulate our problem as: 
\begin{align}
& \text{Pretraining:}~
\min_{\theta}\mathbb{E}_{{x}\in \mathcal{D } }  \ell_{\mathrm{CL}}(x+\delta, x; \theta) \label{eq:pretraining} \\
& \text{Linear finetuning:}~\displaystyle
\min_{\theta_{\mathrm{c}}}\mathbb{E}_{{(x,y)}\in{\mathcal D}}\, \ell_{\mathrm{CE}}( \phi_{ \theta_{\mathrm{c}}}\circ e_{\theta}(x), y)   \label{eq:finetuning}
\end{align}
where $\mathcal{D}$ denotes the training set, $\boldsymbol{\theta}$ represents the parameters of the model. $\boldsymbol{x}$ denotes the original point clouds object. $\ell_{\mathrm{CL}}$ denotes the designed robustness aware contrastive loss with parameter $\boldsymbol{\theta}$, and the adversarial perturbation $\boldsymbol{\delta}$ under budget $\epsilon$. During linear finetuning phase \eqref{eq:finetuning}, $\ell_{\mathrm{CE}}$ is the cross-entropy loss that optimize parameters of linear prediction head $\phi_{\theta_{\mathrm{c}}} $ and $e_{\theta}$ is the fixed robustness aware feature encoder that we obtained after adversarial pretraining stage \eqref{eq:pretraining}. $\phi_{\theta_{\mathrm{c}}} \circ e_{\theta}$ denotes the
classifier by equipping the linear prediction head $\phi_{\theta_{\mathrm{c}}} $
on top of the fixed feature encoder $e_{\theta}$.
\subsection{ Robust Adversarial Contrastive Learning }
\begin{figure}
	\centering
	\includegraphics[width=\linewidth]{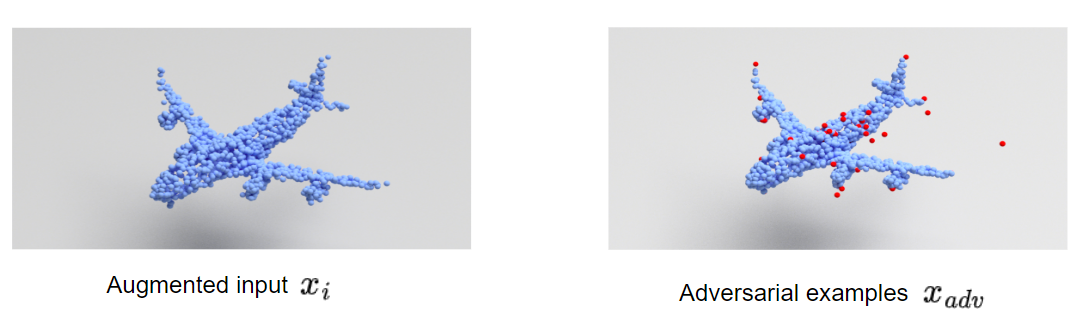}
	\caption{ Visualization of adversarial samples based on augmented input $x_i$, highly perturbed points are marked red}
	\label{fig:compare}
	\vspace{-10pt}
\end{figure}

\paragraph{Adversarial examples} We now introduce our method on how to achieve adversarial robustness of representations in contrastive learning manner. Because the mechanism of adversarial training is to minimize loss of adversarial samples, the first step of our method is to generate label-free adversarial samples during constrastive learning. Fig.\ref{fig:compare} shows the visualization result of one adversarial example and we can see only a few perturbed points are human-noticeable.
\begin{figure}
\removelatexerror
{\centering
\begin{minipage}{1\linewidth}
\begin{algorithm}[H]
\SetAlgoLined
\SetKwInOut{Input}{Input}
\Input{A set of point cloud objects $\boldsymbol{x}$; Augmentation family $\mathcal{T}$; feature encoder $e$; perturbation budget $\epsilon$ ,number of steps $t$}
\SetKwInOut{Result}{Result}
\Result{ adversarial samples $x_{adv}$}
Augment $\boldsymbol{x}$ to be (${\boldsymbol{\tilde{x}_{i}}}$,${\boldsymbol{\tilde{x}_{j}}}$)  with two augmentations sampled from $\mathcal{T}$.\\
Generate a initial small random perturbation $\delta$ and $x_{adv}$=$\tilde{x}_{i}+\delta$

\For {$i$ $\in$  $t$}{
    Generate the corresponding adversarial point clouds with
    $$\boldsymbol{\delta}= \argmax_{{\|\delta\|_{\infty} \leq \epsilon} }KLD(e({\boldsymbol{x}'} ),e({\boldsymbol{x}_{adv}}))$$\\
    $$\boldsymbol{x}_{adv}=\boldsymbol{x}_{adv}+\boldsymbol{\delta}$$
}
\SetKwInOut{return}{return}
\return {$\boldsymbol{x}_{adv}$}
 \caption{Generate adversarial point clouds examples with untargeted I-FGM}
 \label{algo:attacking}
\end{algorithm}
\end{minipage}
\par}
\end{figure}

A number of prior works \cite{acl,advcl,kim2020adversarial} have proposed to use projected representation from self supervised model and constrastive loss to guide attacking algorithm's generation of adversarial samples. Inspired by Virtual Adversarial Training (VAT)\cite{miyato2018virtual}, we introduce a new method that uses Kullback-Leibler divergence (KLD) of unprojected representation between adversarial samples and augmented inputs to calculate gradient direction for updating adversarial perturbation. Following the idea of AT\cite{AT2017}, the loss function in leading attack algorithm can be written as
\begin{eqnarray}
L_{\rm adv}(x,\theta) = Div\left[q(y|x), p(y|x+\delta,\theta)\right] \label{eq:adv_loss}\\
{\rm where}\ \delta = \argmax_{\| \delta\|_\infty\leq\epsilon} Div\left[q(y|x), p(y|x + r, \theta)\right], \label{eq:adv_ptb}
\end{eqnarray}
where $Div[p, p']$ is a non-negative function that measures the divergence  between two distributions $p$ and $p'$. Because the true distribution of the output label, $q(y|x)$, is unknown, we use its \textit{current} estimate $p(y|x,{\theta})$.
The goal of this loss function is to approximate the true distribution $q(y|x)$ by a parametric model $p(y|x, \theta)$ that is robust against adversarial attack to $x$.

Since we do not know the label $y$ in self-supervised training, we rewrite eq:\ref{eq:adv_ptb} by approximating divergence of unprojected representation'distributions between perturbed object $x+\delta$ and augmented original object $x$ in contrastive learning framework. 
\begin{eqnarray}
 L_{\rm adv}(x,\theta) = Div\left[p(y|x,\theta), p(y|x+\delta,\theta)\right] \label{eq:adv_loss2}\\
\delta = \argmax_{\| \delta\|_\infty\leq\epsilon} Div\left[e_{\theta}(x), e_{\theta}(x+\delta)\right], \label{eq:adv_ssl}
\end{eqnarray}
\paragraph{High difference points cloud}
Because high-frequency information is a crucial contributing factor to improving the robustness of perception network in 2D image domain ~\cite{wang2020high,LPF,advcl}. For 3D point clouds, we notice difference between normal of local surface and normal of global surface can reflect the frequency of surface gradient. We propose to use Difference of Normals(DoN)\cite{DON2012} to build a point saliency map according to multi-scale normal estimation and select higher difference points as an additional input during pretraining stage to enhance robustness of our model. 

\begin{figure}[htb]
	\centering
	\includegraphics[width=0.8\linewidth]{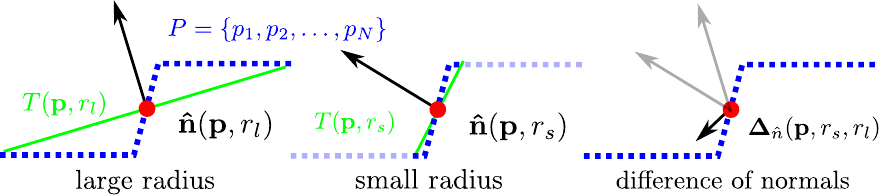}
	\caption{The normal support radius' relation to scale.\cite{DON2012}}
	\label{fig:normalsupporteffect}
\end{figure}

The Difference of Normals (DoN) operator $\mathbf{\Delta}_\mathbf{\hat{n}}$ for any point $\mathbf{p}$ in a point cloud $\xb$, is defined as:
	\begin{align}
		\mathbf{\Delta}_\mathbf{\hat{n}}({p}, r_1, r_2) =& \frac{\mathbf{\hat{n}}({p}, r_1) - \mathbf{\hat{n}}({p}, r_2)}{2},
	\label{eq:don}
	\end{align}
where $r_1, r_2 \in \mathbb{R}$, $r_1<r_2$, and $\mathbf{\hat{n}}(\mathbf{x}, r)$ is the surface normal estimated at point $\mathbf{p}$, given the support radius $r$. For each point clouds object or sense $x$ with number of points $N$, we remove low difference points based on $\mathbf{\Delta}_\mathbf{\hat{n}}({p}, r_1, r_2)$ and keep high difference points with number $c \times N$, where $c \in (0,1)$.
\begin{figure}
	\centering
	\includegraphics[width=\linewidth]{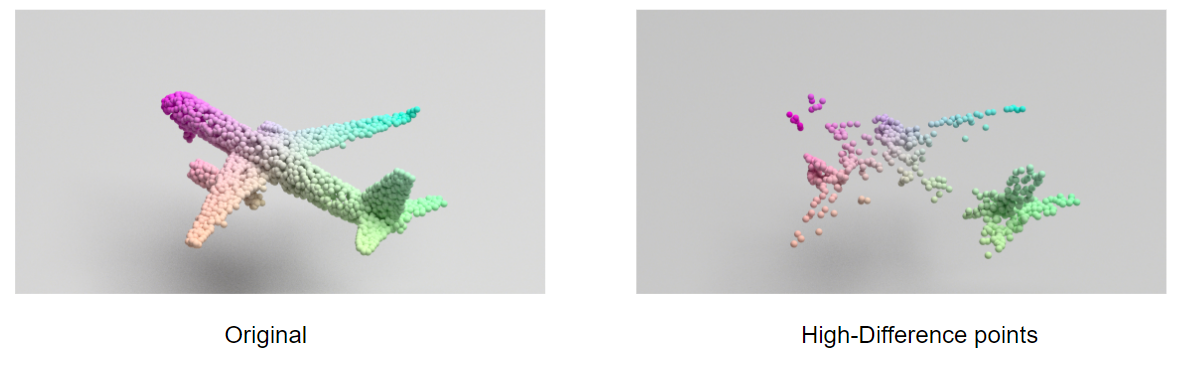}
	\caption{Visualization result for selected points after removing 75\% lower difference points in ModelNet40 }
	\label{fig:HD_img}
\end{figure}
Let $\mathcal{D}$ denote DoN operator, an input point clouds object/scene $X$ can then be decomposed into its high difference part $X_{\mathrm hd}$ and low-difference part$X_{\mathrm ld}$: 
\begin{align}
[X_{\mathrm hd}, X_{\mathrm ld}] = \mathcal D(X).
\end{align}
The motivation is that surface normals estimated at any given radius reflect the underlying geometry of the surface at the scale of the support radius. By calculating the difference of multi-radius estimated surface normals, we can obtain the surface gradient. Thus, we can use DoN to select points which have high frequency information and use them as an additional view in robust contrastive learning eq:\ref{eq: CL_4}. 
\paragraph{Multi-view robust contrastive learning}

We first review the NT-Xent loss used in SimCLR\cite{simclr}. The contrastive loss with a positive augmentation pair $(\tau_{1}(x), \tau_{2}(x))$ from each data input $x$ is given by 
{\begin{align}
& \ell_{\mathrm{CL}}(\tau_1(x), \tau_{2}(x))
=\notag\\ 
&-  \sum_{i=1}^2 \sum_{j \in \mathcal P(i)} \log   \frac{\exp\big(
\text{sim}(z_{i}, z_{j})/t\big)}{\displaystyle \sum_{k\in\mathcal N(i)}  \exp\big(\text{sim}(z_{i}, z_k)/t\big)},
\label{eq: CL_2view}
\end{align}}%

where $z_i = g\circ e(\tau_i(x))$ is the projected feature representation under the $i$th view, $\mathcal N(i)$ represents the set of augmented batch data not including the point $\tau_i(x)$. $\mathcal P(i)$ denote the set of positive views except $i$. $\text{sim}(z_{i1}, z_{i2})$ is the cosine similarity between projected representations $z_{i1}, z_{i2}$ from two views of the same data $x$ and $t$ is a temperature parameter.

For involving adversarial examples and high-difference points as additional inputs, we follow the multi-view contrastive loss in \cite{khosla2020supervised} 
{\begin{align}
& \ell_{\mathrm{CL}}(\tau_1(x), \tau_{2}(x), \ldots, \tau_m(x)) 
=\notag\\
&-  \sum_{i=1}^m \sum_{j \in \mathcal P(i)} \log   \frac{\exp\big(
\text{sim}(z_{i}, z_{j})/t\big)}{\displaystyle \sum_{k\in\mathcal N(i)}  \exp\big(\text{sim}(z_{i}, z_k)/t\big)}, 
\label{eq: CL_mview}
\end{align}}%
where $m$ denotes the number of views as input. By taking selected point clouds $x_{hd}$ from DoN as third view input and adversarial samples $(x+\delta)$ as fourth view input, we propose our contrastive loss function as 
\vspace{-8mm}
{\begin{center}
\begin{equation}
\ell_{\mathrm{CL}}(\tau_1(x), \tau_{2}(x),(x_{hd}), (x+\delta)) 
\label{eq: CL_4}
\end{equation}
\end{center}}%
To minimize the distance between representation from normal input $x$ and other inputs, adversarial samples $x+\delta$ and high difference points $x_{hd}$, we add $\mathcal{KLD}(h_1,h_{adv})$ and $\mathcal{KLD}(h_{adv},h_{hd})$ as the regularization term to our object loss function, where $h_1=e(\tau_1(x))$ and $h_{adv}=e(x+\delta)$. Thus, we can force our backbone network to generate more stable and similar representations for both normal point clouds inputs and adversarial attacking point clouds inputs resulting in downstream task models that are more robust to adversarial samples. Eq:\ref{eq: loss_total} shows our loss funtion for updating parameters.
{\begin{align}
\mathcal{L}_{total}&:=\ell_{\mathrm{CL}}(\tau_1(x), \tau_{2}(x),(x_{hd}), (x+\delta))\notag\\ 
&+\alpha\mathcal{KLD}(h_1,h_{adv})+\beta\mathcal{KLD}(h_{adv},h_{hd})
\label{eq: loss_total}
\end{align}}%
\vspace{-10pt}
\begin{figure}[h]
\removelatexerror
{\centering
\begin{minipage}{1\linewidth}
\begin{algorithm}[H]
\SetAlgoLined
\SetKwInOut{Input}{Input}
\Input{A set of point clouds  $\boldsymbol{x}$,a set of high difference point clouds $\boldsymbol{x_{hd}}$; Augmentation family $\mathcal{T}$; Network backbone and projection head $e$, $g$;}
\SetKwInOut{Result}{Result}
\Result{The  parameters $\boldsymbol{\theta}$ in $e$ and $g$;}
\For{sampled mini-batch $\boldsymbol{x}$}{ 
Augment $\boldsymbol{x}$ to be ($\tilde{\boldsymbol{x}}_{i}$, $\tilde{\boldsymbol{x}}_{j}$) with two augmentations sampled from $\mathcal{T}$. \\ 
Generate the adversarial mini-batch  ($\tilde{\boldsymbol{x}}_{i} +\boldsymbol{\delta}$) with algorithm \ref{algo:attacking} \\
$\ell = \ell_{\mathrm{CL}}(\tilde{\boldsymbol{x}}_{i}, \tilde{\boldsymbol{x}}_{j},(\tilde{\boldsymbol{x}}_{i} +\boldsymbol{\delta}) +\alpha\mathcal{KLD}(h_1,h_{adv}) +\beta\mathcal{KLD}(h_{adv},h_{hd})$ \\
Update parameters ($\boldsymbol{\theta}_{\textit{e}}$, $\boldsymbol{\theta}_{\textit{g}}$) to minimize $\ell$.
}
\caption{Algorithm of Pretraining}
\label{algo:pretraining}
\end{algorithm}
\end{minipage}
\par}
\end{figure}
\vspace{-15pt}
\section{Experimental Results}
\paragraph{Evaluation setting} 
In this section, we validated our adversarial unsupervised learning method on public benchmark point cloud datasets. Specifically, we evaluated our method on two downstream tasks: classification and segmentation. We did not perform model full finetuning since tuning full network weights is not possible for the model to preserve robustness\cite{chen2020}. Instead, we froze the pretrained backbone weights from our self-supervised training to keep the robustness and performed standard finetuning \textbf{(SF)} of prediction heads in different downstream tasks. For attacking algorithm, we chose untarget $\ell_\infty$ I-FGSM\cite{extend_liu} for point clouds with a fixed iteration number and budget $\epsilon$. In 3D classification task, we ran 7 iteration steps and $\epsilon=0.01m$ in attacking during robustness evaluation. In 3D segmentation, we ran 15 iteration steps and $\epsilon=0.08m$ in attacking during robustness evaluation.
\paragraph{Pre-training}
For the backbone pretraining stage, we use the Adam optimizer with a cosine decay learning rate schedule, the exponential moving average parameter starts with $\tau_{start}= 0.996$ and is gradually increased to $1$ during the training. For 3D object classification task, we train our PointNet backbone network in 50 epochs with learning rate 0.001 over 256 batch size and the number of input is 2048. For 3D object segmentation task, we train our DGCNN backbone network in 100 epochs with learning rate 0.0002 over 32 batch size and the number of input is 4096.

We used a combination of the following augmentation method to construct the augmentation family as in \cite{huang2021spatio}:
Random rotation, Random translation, Random scaling, Random cropping, Random cutout, Random jittering, Random drop-out, Down-sampling, Normalization.

\begin{itemize}[itemsep=0.1pt]
    \item Random rotation: For each axis, we draw random angles within $15^\circ$ and rotate around it. 
    \item Random translation: We translate the point cloud globally within 10\% of the point cloud dimension.
    \item Random scaling: We scale the point cloud with a factor $s\in[0.8, 1.25]$.
    \item Random cropping: A random 3D cuboid patch is cropped with a volume uniformly sampled between 60\% and 100\% of the original point cloud. The aspect ratio is controlled within $[0.75, 1.33]$.
    \item Random cutout: A random 3D cuboid is cut out. Each dimension of the 3D cuboid is within $[0.1, 0.4]$ of the original dimension. 
    \item Random jittering: Each point's 3D locations are shifted by a uniformly random offset within $[0, 0.05]$. 
    \item Random drop-out: We randomly drop out 3D points by a drop-out ratio within $[0, 0.7]$.
    \item Down-sampling: We down-sample point clouds based on the encoder's input dimension by randomly picking the necessary amount of 3D points.
    \item Normalization: We normalize the point cloud to fit a unit sphere while training on synthetic data.
\end{itemize}

\subsection{3D Object Classification}\label{sec:classification}

We first verified our proposed method on classification task. In this downstream task,we used PointNet \cite{pointnet} with SimCLR\cite{simclr} as backbone. During evaluation, we apple two different settings: standard finetune and adversarial full finetune (AFF). For standard finetune, we add a linear layer to the representation obtained from the pretrained backbone in self-supervised training and only finetune the linear layer with training data. For adversarial full finetune (AFF), we first use attacking algorithm I-FGSM\cite{extend_liu} to generate an adversarial example for each input of training data in a supervised manner. Then, we add a linear layer and we finetune the whole network including the pretrained backbone with original training data and adversarial examples.

We trained the backbone network on ModelNet40\cite{modelnet} dataset with $\alpha=1,\beta=1$, and evaluated downstream tasks on ModelNet40\cite{modelnet} dataset. For each point cloud object we selected 2048 points as input with only coordinates and removed 512 low difference points with DoN operator as HD input. 

Experimental results are shown in Table.\ref{tab:classification}. For standard finetune, our pretraining model PointACL achieved 27.51\% robust accuracy under I-FGSM attack in testing set of ModelNet40, which improved \textbf{23.48\%} over baseline pretraining model SimCLR. We also had more than \textbf{14\%} robust accuracy improvement over other adversarial contrastive learning methods. 

\begin{table}[h]
  \centering
  \caption{\textbf{Performance result of different methods in evaluation.} Standard Accuracy(\textbf{SA}) represent the accuracy in test dataset and Robust Accuracy (\textbf{RA}) means the accuracy evaluated under test dataset generated by  untarget I-FGSM\cite{extend_liu} for point clouds with $\epsilon$=0.01m in 7 steps  }
  \scalebox{0.55}{
  \begin{tabular}{@{}lcccc@{}}
    \toprule
     Training type & Method & Standard Accuracy(\%) & Robust Accuracy(\%) \\
    \midrule
    \multirow{1}{*}{Supervised}& AT\cite{AT2017} & 82.54   & 44.49 \\
    \midrule
    \multirow{4}{*}{Self-supervised+finetune} & SimCLR\cite{simclr} & 86.33 & 4.03 \\
    &{}RoCL\cite{kim2020adversarial} & 85.22 & 8.72 \\
    &{}ACL\cite{acl} & 85.85 & 10.25 \\
    &{}PointACL(Ours) w/o HD & 82.28 & \bf24.26 \\
    &{}PointACL(Ours) & 80.71 & \bf27.51 \\
    \bottomrule
  \end{tabular}
  }
  \label{tab:classification}
\end{table}

\subsection{3D Sementic Segmentation}

In 3D Segmentic Segmentation, we used DGCNN \cite{dgcnn} with SimCLR\cite{simclr} structure as backbone. During evaluation, we added 2-layered MLP network to the representation obtained from the pretrained backbone in self-supervised training for segmentataion. We trained the bakcbone network on S3DIS\cite{S3DIS} dataset on area 1-5 with $\alpha=1,\beta=1$, and evaluated on area 6. We selected 4096 points as input with only coordinates and removed 1024 lower difference points with DoN eq:\ref{eq:don} as HD input.

As Table.\ref{tab:segmentation} shows, our pretraining model PointACL increased \textbf{25.23\%} robust accuracy and \textbf{13.52\%} mIoU in robustness evaluation over baseline model SimCLR. Compared to other adversarial contrastive learning methods, our method had more than \textbf{15.13\%} robust accuracy with RoCL\cite{kim2020adversarial} and more than \textbf{12.26\%} robust accuracy with ACL\cite{acl}.

\begin{table}[h]
  \centering
  \caption{\textbf{Performance result of different methods in evaluation.} Standard Accuracy(\textbf{SA}) represent the accuracy and (\textbf{S-mIoU}) means the standard mean IoU in area 6 and Robust Accuracy (\textbf{RA})/\textbf{R-mIoU} represents the accuracy/IoU evaluated under test dataset generated by  untarget I-FGSM\cite{extend_liu} for point clouds with $\epsilon$=0.08m in 15 steps  }
  \scalebox{0.55}{
  \begin{tabular}{@{}lccccc@{}}
    \toprule
     Training type & Method & SA(\%) & S-mIoU(\%)& RA(\%)& R-mIoU(\%) \\
    \midrule
    \multirow{1}{*}{Supervised}& AT\cite{AT2017} & 80.29&52.99&55.61&28.53   \\
    \midrule
    \multirow{4}{*}{Self-supervised+finetune} & SimCLR\cite{simclr} & 82.37 &55.54 &13.85& 5.60 \\
    &{}RoCL\cite{kim2020adversarial} & 79.53 & 52.06 &23.95&11.37 \\
    &{}ACL\cite{acl} & 78.10 & 49.79&26.82&12.41 \\
    &{}PointACL(Ours) w/o HD & 79.06 &50.27&\bf36.16& \bf19.01 \\
    &{}PointACL(Ours)  & 78.69 &49.85&\bf39.08& \bf19.12 \\
    \bottomrule
  \end{tabular}
  }
  \label{tab:segmentation}
\end{table}

\subsection{Tradeoff between Robust Accuracy (\textbf{RA}) and Standard Accuracy(\textbf{SA})}\label{sec:tradeoff}

\begin{figure}[h]
    \vspace{-10pt}
	\centering
	\includegraphics[width=\linewidth]{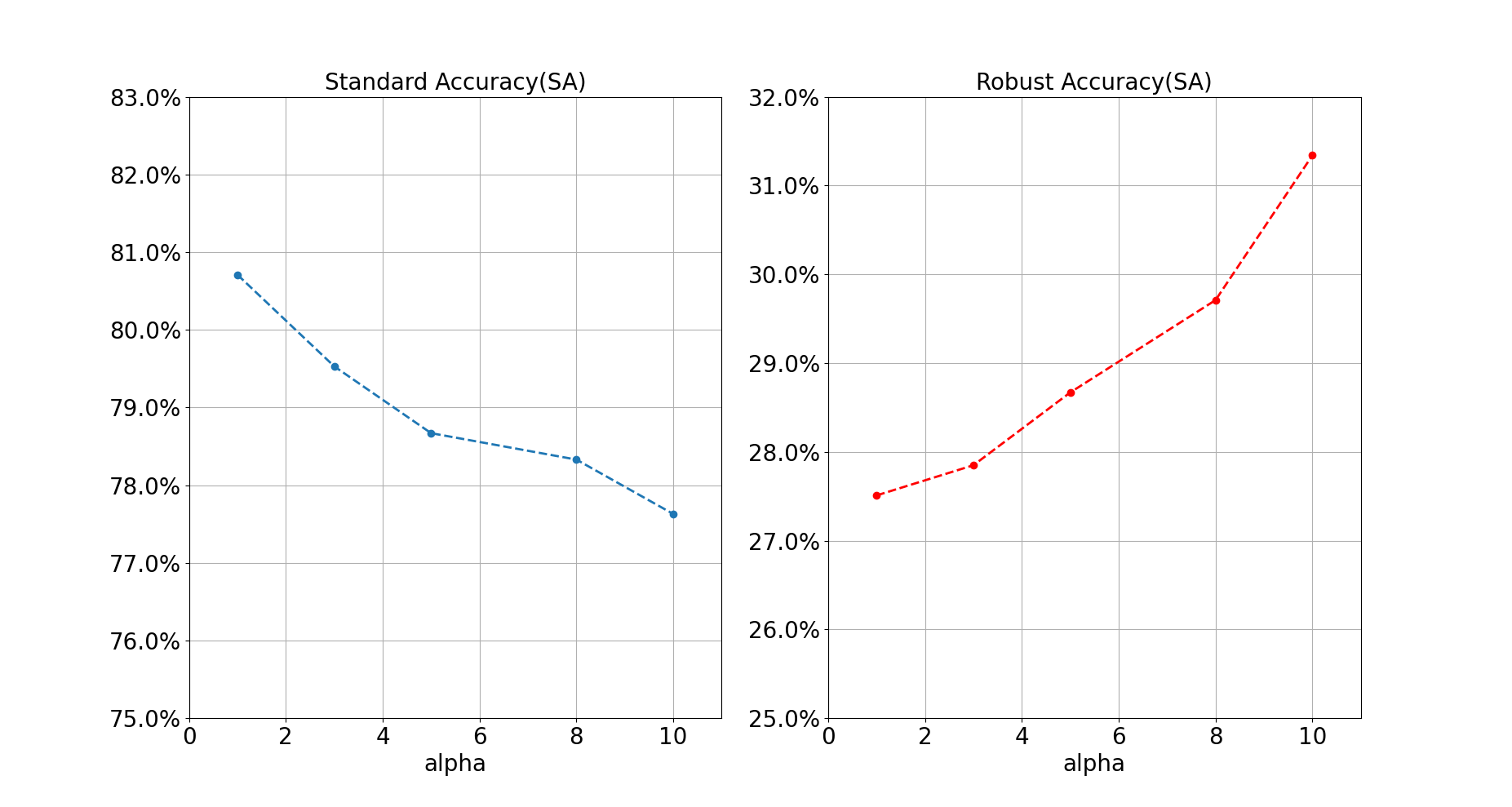}
	\caption{Standard Accuracy(SA) and Robust Accuracy(RA) with different $\alpha$ }
	\label{fig:tradeoff}
\end{figure}
We found that tuning the hyperparameter $\alpha$ yielded increasing peformance of robust accuracy (\textbf{RA}) at the cost of decreasing standard accuracy(\textbf{SA}). Fig.\ref{fig:tradeoff} shows the \textbf{RA} and \textbf{SA} under different $\alpha$ in 3D object classification task. One possible reason is that increasing $\alpha$ forces the backbone network to generate similar unprojected representation ${h}$ between adversarial sample and normal point clouds. This helps the model make more accurate predictions when facing attacking inputs but misleads it when the input data is clean.

\subsection{Robustness evaluation vs. attack
strength}
\begin{figure}[h]
	\centering
	\includegraphics[width=1.1\linewidth]{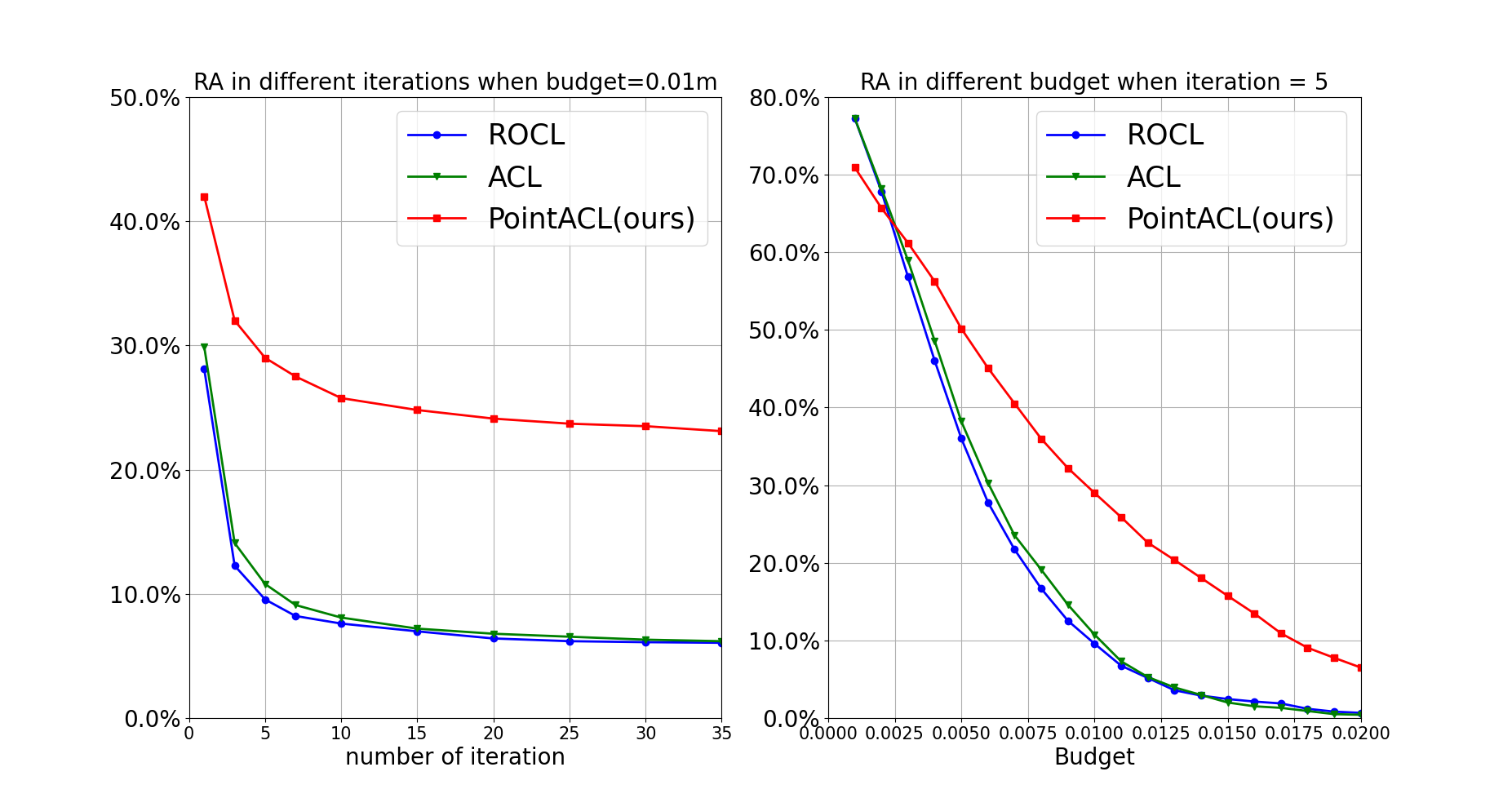}
	\caption{Robust accuracy for 3D classification in ModelNet with different steps and budget. \textbf{Left:} We fix attacking budget=0.01m and set different iteration.\textbf{Right:} We fix iteration=5 and set different attacking budget.}
	\label{fig:attack_strength}
	\vspace{-8pt}
\end{figure}

The strength of attacking algorithm I-FGM\cite{IFGM} are affected by two things:(a) The number of iteration steps; increasing the number of iterations produced strong adversarial samples but increasing the number of steps did not generate stronger adversarial samples\cite{athalye2018obfuscated}. The left graph of Figure.\ref{fig:attack_strength} shows that attacking strength slowed after iteration 5 and stopped increasing after iteration 25. Our method outperformed other methods in robust accuracy at different number of iterations; (b) The attacking budget(m) $\epsilon$, which sets perturbation boundary of attacking, is also very important for attacking algorithm; increasing $\epsilon$ significantly enhances the attacking strength. We tested all the pretraining methods with budget from 0.001m to 0.02m with iteration 5. We found the performance of ROCL and ACL to be nearly 0\% when $\epsilon>0.02m$. Our method was better when $\epsilon>0.002m$.

\subsection{Robustness transferability across datasets}
In table.\ref{tab:cross_dataset}, we evaluated the transferability of model robustness across different datasets for downstream 3D classification task. Following the same setting from Section. \ref{sec:classification}, where $A \to B$ denotes the transferability from pretraining on dataset $A$ to finetuning on another dataset $B$ ($\neq A$). We chose ShapeNetCore\cite{shapenet2015} as our secondary dataset. It contained over 37000 3D point cloud objects with 55 categories in the training set while ModelNet40 only had around 9840 objects. In table.\ref{tab:cross_dataset}, ${ShapeNetCore\xrightarrow[]{}ModelNet40}$ scenario had very stable standard accuracy for all pretraining methods. Our method's robust accuracy was 13.57\% higher than baseline model and 8.18\% higher than RoCL when $\alpha$=10. In ${ModelNet40 \xrightarrow[]{} ShapeNetCore}$ scenario, our method significantly improved the robustness performance over baseline and other methods while suffering only a small drop to standard accuracy. This pattern of large RA gain and small SA drop was also observed in \ref{sec:tradeoff}.  
\begin{table}[h]
  \centering
  \caption{\textbf{Performance result of different methods in cross dataset evaluation between ModelNet40 and ShapeNetCore.} ModelNet40 $\xrightarrow[]{}$ ShapeNetCore means we pretrained the model on ModelNet40 and linear evaluate on ShapeNetCore dataset. The robust accuracy evaluated under test dataset generated by  untarget I-FGSM\cite{extend_liu} for point clouds with $\epsilon$=0.01m in 7 steps  }
  \scalebox{0.65}{
  \begin{tabular}{@{}ccc|cc@{}}
    \toprule
     {} & \multicolumn{2}{c|}{ModelNet40 $\xrightarrow[]{}$ ShapeNetCore} & \multicolumn{2}{c|}{ShapeNetCore$\xrightarrow[]{}$ ModelNet40} \\
    \toprule
      Pretraing method & SA(\%) & RA(\%)& SA(\%) & RA(\%) \\
    \midrule
    PointNet (SimCLR)\cite{simclr} & 81.10 &2.13 &85.49& 5.39 \\
    RoCL\cite{kim2020adversarial} & 81.29 & 2.46 &85.01&10.78
 \\
    ACL\cite{acl} & 81.93 & 2.99 & 85.29&16.09 \\
    PointACL(Ours,$\alpha$=1) & 80.01 &\bf13.34&85.04 & 10.56 \\
    PointACL(Ours,$\alpha$=10) & 77.16 &\bf26.12&84.56& \bf18.96 \\
    \bottomrule
  \end{tabular}
  }
  \label{tab:cross_dataset}
\end{table}
\vspace{-20pt}

\section{Ablation Study}

\subsection{Projected Representation vs Unprojected Representation}
In SimCLR\cite{simclr}, the authors showed that a nonlinear projection head improves the representation quality for contrastive learning. A number of prior works \cite{kim2020adversarial,acl,advcl} used contrastive loss with projected feature representation ${\boldsymbol{z}}$ to generate adversarial samples during adversarial contrastive learning. In our pretraining method, we achieved better robustness performance (Table.\ref{tab:representation compare}) using unprojected feature representation ${\boldsymbol{h}}$. To further evaluate this approach, we also experimented with using unprojected feature representation ${\boldsymbol{h}}$ instead of ${\boldsymbol{z}}$ to calculate contrastive loss in generating adversarial samples for RoCL\cite{kim2020adversarial} and ACL\cite{acl}. Because we wanted to focus only on the importance of feature representation selection during adversarial contrastive learning, we didn't include high difference(HD) point clouds input in this experiment. 

The results in Table.\ref{tab:representation compare} showed that using unprojected feature representation ${\boldsymbol{h}}$ resulted in better robustness performance in our method but not in other adversarial contrastive learning methods. If we look at the regularization term $\mathcal{KLD}(h_1,h_{adv})$ in our pretraining loss function, we find an explanation for this. The regularization improves the model's robustness by minimizing the distance between ${\boldsymbol{h_1}}$ and ${\boldsymbol{h_{adv}}}$ (unprojected representation). Because the prediction result from robustness testing on downstream tasks is based on unprojected representation ${\boldsymbol{h}}$ and not projected representation ${\boldsymbol{z}}$, replacing ${\boldsymbol{h}}$ with ${\boldsymbol{z}}$ in our method will decrease the robustness performance of the model during adversarial contrastive training.

\begin{table}[h]
  \centering
  \caption{\textbf{3D classification performance under ModelNet40 with different representation choice } The robust accuracy evaluated under test dataset generated by  untarget I-FGSM\cite{extend_liu} for point clouds with $\epsilon$=0.01m in 7 steps }
  \scalebox{0.65}{
  \begin{tabular}{@{}l|c|c|c@{}}
    \toprule
      Pretraining method & Standard Accuracy(\%) & Robust Accuracy(\%) \\
    \midrule
    SimCLR\cite{simclr} & 86.33 & 4.03 \\
    RoCL\cite{kim2020adversarial} & 85.22 & \bf8.72 \\
    RoCL\cite{kim2020adversarial}(use $\boldsymbol{h}$) & 85.98 & 5.83 \\
    ACL\cite{acl} & 85.85 & \bf10.25 \\
    ACL\cite{acl}(use $\boldsymbol{h}$) & 86.46 & 6.24 \\
    PointACL(Ours)(use $\boldsymbol{z}$) w/o HD & 85.62 & 8.72 \\
    PointACL(Ours)(use $\boldsymbol{h}$) w/o HD & 82.28 & \bf24.26 \\
    
    \bottomrule
  \end{tabular}
  }
  \label{tab:representation compare}
\end{table}
\subsection{Loss function analysis }
To better understand the importance of $\mathcal{KLD}(h_1,h_{adv})$ and $\mathcal{KLD}(h_{adv},h_{hd})$ in our pretraining method, we performed the following 3D classification experiment in ModelNet. From Table.\ref{tab:loss_ab}, we observed that the model has 13.83\% robust accuracy advantage when we set ($\alpha=1,\beta=0$) compare with the  baseline model ($\alpha=0,\beta=0$). From that result, we can find $\mathcal{KLD}(h_1,h_{adv})$ part plays a important role in improving robustness to our pretrained model. When we use setting ($\alpha=1,\beta=1$), we can see $\mathcal{KLD}(h_{adv},h_{hd})$ part in our loss function further increased robustness, which proves the contribution of High-difference points.

\begin{table}[h]
  \centering
  \caption{\textbf{Ablation study of PointACL }}
  \scalebox{0.7}{
  \begin{tabular}{@{}l|c|c|c@{}}
    \toprule
      Pretraining method & Standard Accuracy(\%) & Robust Accuracy(\%) \\
    \midrule
    PointACL($\alpha=0,\beta=0$) & 86.79 & 11.94 \\
    PointACL($\alpha=1,\beta=0$) & 82.02 & 25.77 \\
    PointACL($\alpha=1,\beta=1$) & 80.71 & 27.51 \\
    
    \bottomrule
  \end{tabular}
  }
  \label{tab:loss_ab}
\end{table}
\vspace{-10pt}
\section{Conclusion}
In this paper, we have studied methods to make contrastive learning pretrained model more robust in the 3D point clouds domain. We have showed that using virtual adversarial loss to generate adversarial samples are  beneficial towards robustness. We have further showed that using difference of normal (DoN) operator to select high difference points as additional input view can enhance the robustness. 
Our proposed approaches can achieve state-of-the-art robust accuracy using standard linear finetuning in two downstream tasks: 3D object classification and 3D segmentic segmentation. Extensive experiments involving cross-datasets and attacking strength have also been made to demonstrate universality of our method in improving robustness.
\newpage
{\small
\bibliographystyle{ieee_fullname}
\bibliography{egbib}
}

\end{document}